\documentclass[11pt]{article}
\usepackage{epic,latexsym,amssymb}
\usepackage{amsfonts}
\usepackage{amscd}
\usepackage{amsmath}
\usepackage{graphicx}
\usepackage{color}
\usepackage{caption,
caption}
\usepackage{tikz}
\usepackage{amsthm}
\usepackage{bbm}
\usepackage{amsfonts}

\usepackage{caption}
\usepackage[margin=0.25in]{geometry}
\usepackage{pgfplots}
\pgfplotsset{width=10cm,compat=1.9}

\usepackage{epic,latexsym,amssymb}
\usepackage{tikz}
\usepackage[bold,full]{complexity}
\usepackage{tikz,epsf}
\usepackage{hyperref}
\usepackage{graphicx}[width=\textwidth]
\usepackage{amsmath}
\usepackage{listings}

\usepackage{amsfonts}
\usepackage{amscd}
\usepackage{amsmath}
\usepackage{graphicx}
\usepackage{color}
\usepackage{caption,subcaption}
\usepackage{lineno}

\usepackage{graphicx}

\usepackage[ruled,vlined]{algorithm2e}

\usepackage{lineno}

\newtheorem{thm}{Theorem}
\newtheorem{conj}[thm]{Conjecture}

\newcommand{\QEDmark}{\mbox{\textsc{qed}}}
\newcommand{\proofStarter}[1]{\textsc{#1} }

\def\vertex(#1){\put(#1){\circle*{2}}}
\def\vertexo(#1){\put(#1){\circle{2}}}
\def\vert(#1){\put(#1){\circle*{1.5}}}
\def\verto(#1){\put(#1){\circle{1.5}}}
\def\lab(#1)#2{\put(#1){\makebox(0,0)[c]{#2}}}
\definecolor{DarkGreen}{rgb}{0.2, 0.6, 0.3}

\definecolor{electricindigo}{rgb}{0.44, 0.0, 1.0}

\let\oldenumerate\enumerate
\renewcommand{\enumerate}{
  \oldenumerate
  \setlength{\itemsep}{0.5pt}
  \setlength{\parskip}{0pt}
  \setlength{\parsep}{0pt}
}

\begin{document}

\title{Artificial intelligence and machine learning generated \\ conjectures with TxGraffiti}
\author{$^{1,2}$Randy Davila\
\\
$^1$Research and Development \\
RelationalAI \\
Berkeley, CA 94704, USA\\
\small {\tt Email: randy.davila@relational.ai}\\
\\
$^2$Department of Computational Applied \\ Mathematics \& Operations Research\\
Rice University\\
Houston, TX 77005, USA \\
\small {\tt Email: randy.r.davila@rice.edu} \\
\\
}

\date{}
\maketitle

\abstract{\emph{TxGraffiti} is a machine learning and heuristic based artificial intelligence designed to automate the task of conjecturing in mathematics. Since its inception, TxGraffiti has generated many surprising conjectures leading to publication in respectable mathematical journals. In this paper we outline the machine learning and heuristic techniques implemented by TxGraffiti. We also recall its contributions to the mathematical literature and announce a new online version of the program available for anyone curious to explore conjectures in graph theory. }
\\


\section{Introduction}
The ability of carefully designed computer programs to generate meaningful mathematical conjectures has been demonstrated since the late 1980s, notably by Fajtlowicz's GRAFFITI program~\cite{GRAFFITI1}. Indeed, this heuristic-based program was the first artificial intelligence to make significant conjectures in matrices, number theory, and graph theory, attracting the attention of renowned mathematicians like Paul Erdős, Ronald Graham, and Odile Favaron. Inspired by the pioneering work of Fajtlowicz, and by interactions with mathematicians who considered conjectures of GRAFFITI, we developed the \emph{TxGraffiti} program, a modern conjecturing artificial intelligence named in homage to this rich history of conjectures made by GRAFFITI and now available as an \href{https://txgraffiti.streamlit.app}{interactive website}. While our program TxGraffiti draws inspiration from GRAFFITI and its successor Graffiti.pc by DeLaViña~\cite{Graffitipc}, it was developed independently and features several distinct design elements and conjecturing capabilities, which we detail in this paper. 

When discussing computer-assisted conjecturing, we remark that it is easy for a computer to generate many plausible conjectures. For example, one might gather a set of mathematical objects and test various functions applied to these objects to identify potential relationships (inequalities). If a relationship holds across all objects in the database, it becomes a plausible conjecture. For instance, given a database of graphs and the ability to compute various parameters on said graphs, a computer might quickly discover the relation:
\begin{equation}\label{eq:example}
\alpha(G) \leq n(G),
\end{equation}
where $\alpha(G)$ is the \emph{independence number} (the cardinality of a maximum set of pairwise non-adjacent vertices in $G$) and $n(G)$ is the \emph{order} (the number of vertices in $G$). A more refined bound for $\alpha(G)$ in nontrivial, connected graphs is:
\begin{equation}\label{eq:example2}
\alpha(G) \leq n(G) - 1.
\end{equation}
Both inequalities~\ref{eq:example} and \ref{eq:example2} hold under specific conditions, and TxGraffiti is designed to consider various hypotheses to form such conjectures. The mechanism for considering different hypotheses is a heuristic called \emph{Theo}, detailed in Section~\ref{sec:methods}.

To discover relationships like inequalities~\ref{eq:example} and \ref{eq:example2}, TxGraffiti employs a machine learning, data-driven approach using linear optimization methods. This approach allows the program to find optimal parameters \( m, b \in \mathbb{R} \) for example conjecturing on \(\alpha(G)\) in terms of another graph invariant, say \(i(G)\), presenting conjectures in the form:
\begin{conj}
If \(G\) satisfies certain boolean conditions, then
\[
\alpha(G) \leq m \cdot i(G) + b,
\]
where this bound is sharp.
\end{conj}

By integrating machine learning techniques with the Theo and \emph{Dalmation} heuristics (see Section~\ref{sec:related-work}), TxGraffiti generates novel conjectures suitable for publication in mathematical journals. In Section~\ref{sec:related-work}, we discuss the historical development of AI-assisted conjecturing and relevant techniques. Section~\ref{sec:methods} details TxGraffiti's implementation, Section~\ref{sec:results} presents conjectures produced by TxGraffiti that have led to mathematical publications, and Section~\ref{sec:conclusion} provides concluding remarks.


\section{Related Work}\label{sec:related-work}
In 1948, Turing proposed that intelligent machines could be a significant asset in mathematical research, requiring substantial intelligence while involving \emph{``minimal interaction with the external world''}~\cite{Turing}. Following this vision, early work in computer-assisted mathematics includes Newell and Simon's \emph{Logic Theorist} program developed in the 1950s. This program was capable of proving some theorems in first-order logic, and they boldly predicted that a computer would eventually discover and prove a crucial mathematical theorem~\cite{SimonNewell}. This program, among others, focused heavily on theorem proving, including a notable achievement in 1996 being the computer proof of the Robbins Conjecture~\cite{Robbin}. 

Artificial intelligent conjecture-making is the other side of computer-assisted mathematics and began with Wang's work in the late 1950s \cite{Wang}. Wang's \emph{Program II} generated numerous statements in propositional logic, which could be considered conjectures or potential theorems. Despite the program's innovative approach, it struggled to filter the vast number of generated statements to identify those of significant interest, highlighting a key challenge in automated conjecture-making. Indeed, as mentioned prior, a computer may easily generate thousands of plausible relationships on a given database of mathematical objects.  

A breakthrough in computer assisted conjecture-making was achieved with Fajtlowicz’s \emph{GRAFITTI} program, the first to produce conjectures that led to published mathematical research \cite{Graffiti, GRAFFITI1}. Early versions of Graffiti faced the "Sorcerer’s Apprentice Problem," where the challenge was to manage the overwhelming number of generated conjectures. This issue was mitigated by Fajtlowicz’s \emph{Dalmatian} heuristic, which limited both the quantity and ensured the quality of the output conjectures~\cite{Larson}. This heuristic ensured that each conjecture was significant concerning at least one object in the program's database, and by doing so, also removed any new potential conjecture that followed by transitivity from another potential conjecture. 

GRAFITTI and its successor, \emph{Graffiti.pc}, developed by DeLaVina, have been instrumental in advancing computer assisted conjecture-making and have resulted in numerous publications \cite{Graffitipc}. These programs, although not widely distributed, have paved the way for modern automated conjecturing systems. TxGraffiti, while inspired by these programs, was developed independently and incorporates unique design elements and conjecturing capabilities. Namely, TxGraffiti employs a machine learning, data-driven approach using linear optimization methods to generate plausible conjectured inequalities, thereafter, implementing two heuristics for filtering conjectures found; details to be described in Section~\ref{sec:methods}.

Other notable programs in the domain of automated conjecturing include Lenat’s \emph{AM}~\cite{Lenat_1, Lenat_2, Lenat_3}, Epstein’s \emph{GT}~\cite{Epstein_1, Epstein_2}, Colton’s \emph{HR}~\cite{Colton_1, Colton_2, Colton_3}, Hansen and Caporossi’s~\emph{AGX} \cite{AGX_1, AGX_2}, and Mélot’s \emph{Graphedron}~\cite{graphedron_1}. These programs have significantly contributed to various mathematical domains and have demonstrated the potential of automated systems in aiding mathematical discovery.

\section{Methodology}\label{sec:methods}

In designing a computer program that generates mathematical conjectures, the first requirement is a database of mathematical objects. In the case of TxGraffiti, these objects are edge lists of simple connected graphs. It is crucial to underscore the importance of data quality. An extensive database is optional for the computer to identify non-trivial relationships among object properties. Instead, what is needed is a collection of unique instances of the objects in question, such as special counter-examples or interesting families of graphs from the literature. In our implementations, we utilized databases of several hundred objects, though we have also experimented with thousands with little to no meaningful returns in conjecture quality. 

\subsection{Feature Generation}\label{subsec:feature}
\begin{figure}[h]
    \centering
    \includegraphics[width=11.5cm, height=3.5cm]{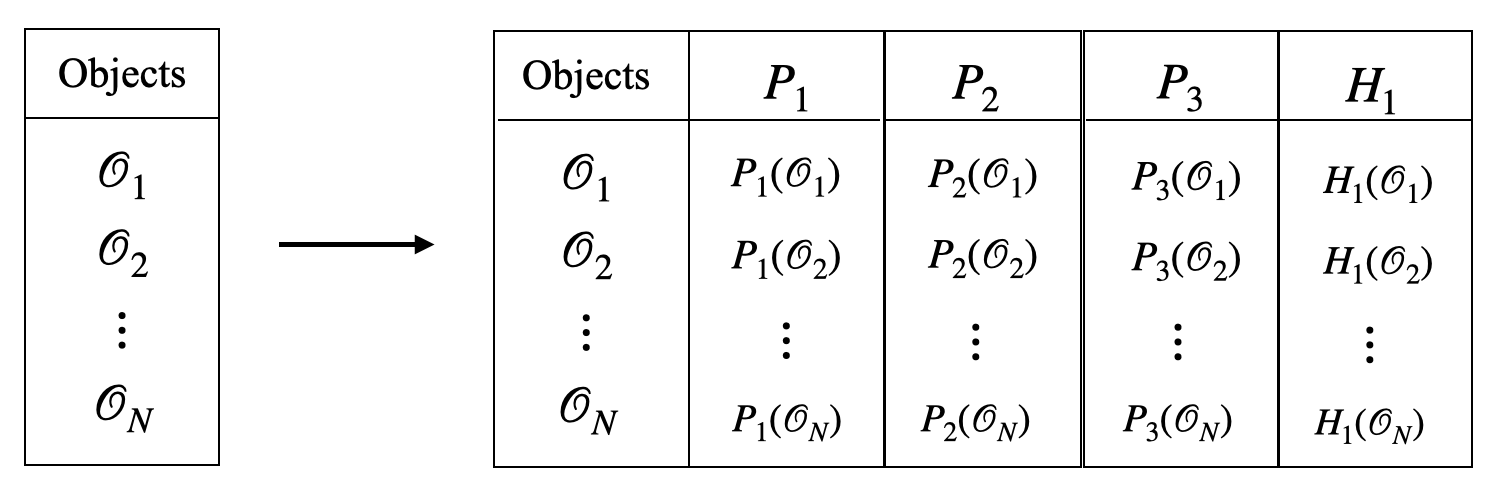}
    \caption{A mapping of a collection of $N$ mathematical objects to a table of numerical and Boolean properties.}
    \label{fig:1}
\end{figure}

After a collection of mathematical objects is collected, the next step in the design of TxGraffiti is to generate a table (a csv file representing a database) of various precomputed functions on the objects in this database. Our framework mandates that at least two of these functions return numerical values (for pairwise comparison), while others can return numerical or Boolean values. See Figure~\ref{fig:1} for an illustration of this process; numerical properties are denoted by $P_i$, and Boolean properties are denoted by $H_i$. 

Indeed, once a table of data like the one is available, TxGraffiti may conjecture on the data; regardless if the data represents graph object data. Thus, if one were to generate conjectures on different types of data, one would first create such a table and then implement the steps in the following subsections. 

\subsection{Inequality Generation}\label{subsec:inequality}
In this section, we propose and implement a simplified version of the following steps for a computer program to generate inequalities relating properties of the objects under consideration, with an emphasis on conjecture simplicity and on conjecture strength. 
\begin{enumerate}
    \item[1.] Select a target property $P_i$ -- a precomputed and numerically valued function on the objects in the database. 
    \item[2.] Choose an inequality direction (upper or lower) to bound the property $P_i$.
    \item[3.] For each precomputed numerical function $P_j$, with $j \neq i$, use a supervised machine learning technique or linear program to find a function $f$ such that $P_i(\mathcal{O}) \le f((P_j(\mathcal{O}))$ holds for each object $\mathcal{O}$ in the database, \emph{and} the number of instances where the inequality is an equality is maximized. 
    \item[4.] If $P_i(\mathcal{O}) \neq  f((P_j(\mathcal{O}))$ for all objects $\mathcal{O}$ in the database, disregard $f$ as a conjectured upper (or lower) bound on $P_i$. Otherwise, $f$ is called a \emph{sharp bounding function}; store $f(P_j)$ as a conjectured upper (or lower) bound on $P_i$ and record the set of objects $\mathcal{O}$ where  $P_i(\mathcal{O}) =  f(P_j(\mathcal{O}))$; the size of this set is the \emph{touch number} of the conjecture. 
\end{enumerate}

\begin{figure}
\centering
\begin{tikzpicture}

\draw[->] (0,0) -- (0,6) node[anchor=south] {$\text{Target } P_i$};
\draw[->] (0,0) -- (6,0) node[anchor=west] {$P_{j \neq i}$};

\draw[dotted, thick] (0,1) -- (5,6);

\foreach \x/\y in {1.2/2, 1.5/2.5, 2/3, 2.5/3.5, 3/4, 3.5/4.5, 1/1.5, 1.8/1.8, 2.2/2.8, 2.7/3.1, 3.3/3.6, 3.7/3.9, 4.2/4.2, 
                   0.5/0.8, 0.8/1.1, 1/1.3, 1.3/1.6, 1.6/1.8, 1.9/2.1, 2.1/2.3, 2.4/2.5, 2.7/2.7, 3/3, 3.2/3.2, 3.5/3.4, 
                   3.8/3.6, 4/3.8, 4.3/4, 4.6/4.2, 4.9/4.5} {
    \fill[green] (\x,\y) circle (3pt);
}

\node[anchor=west] at (4.75,5.5) {Sharp bounding function};
\node[anchor=west] at (4.5,5) {$y = mP_j + b$};

\node[anchor=west] at (5.25,2.75) {\textbf{CONJECTURE:}};
\node[align=left] at (7,2) {
    If $\mathcal{O}$ is an object, then $P_i(\mathcal{O}) \leq mP_j(\mathcal{O}) + b$,\\
    and this bound is sharp.
};

\end{tikzpicture}
\caption{Finding a possible (linear) upper bound on the target property $P_i$ in terms of property $P_j$.}
\label{fig:main1}
\end{figure}
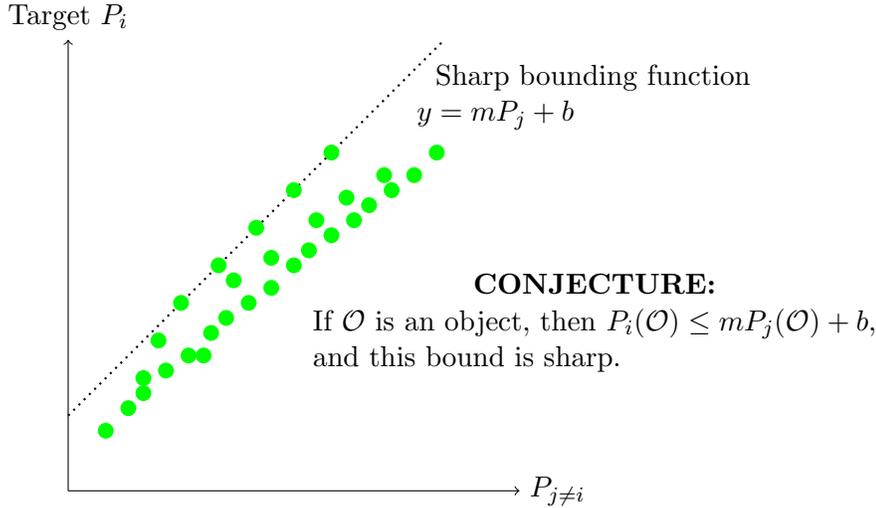

TxGraffiti follows these steps at each instance of a user requesting a desired conjecture, with conjectured upper and lower bounds computed automatically through linear programming formulations. For example, consider producing one conjectured upper bound on $P_i(\mathcal{O})$ in terms of another numerical valued function $P_j(\mathcal{O})$). This is achieved by solving a linear optimization problem, and in the simplest case, TxGraffiti aims to minimize a some linear function $f(m, b)$ subject to a set of constraints.  
\begin{equation*}
\begin{aligned}
& \underset{m, b}{\text{minimize}}
& & f(m, b) \\
& \text{subject to}
& & P_i(\mathcal{O}) \leq mP_j(\mathcal{O}) + b, & \forall \mathcal{O} \in \text{Database},
\end{aligned}
\end{equation*}

The goal is to find the line with the slope $m$ and y-intercept $b$ that satisfies all the inequalities and maximizes the number of times these inequalities hold with equality. In this way, TxGraffiti searches for the best linear upper bound on $P_i(\mathcal{O})$ in terms of $P_j(\mathcal{O})$ that holds for all objects $\mathcal{O}$ in the database given some hypothesis for the objects to satisfy; see Figure~\ref{fig:main1} for a graphical illustration of the linear upper bound.

\begin{figure}
\centering
\begin{tikzpicture}

\draw[->] (0,0) -- (0,6) node[anchor=south] {$\text{Target } P_i$};
\draw[->] (0,0) -- (6,0) node[anchor=west] {$P_{j \neq i}$};

\draw[dotted, thick] (0,1.25) -- (5,4.75);

\foreach \x/\y in {1.2/2, 1.5/2.5, 2/3, 2.5/3.5, 3/4, 3.5/4.5, 1/1.5, 1.8/1.8, 2.2/2.8, 2.7/3.1, 3.3/3.6, 3.7/3.9} {
    \fill[green] (\x,\y) circle (3pt);
}

\foreach \x/\y in {2.2/2.8, 2.7/3.1, 3.3/3.6, 3.7/3.9, 4.2/4.2, 
                   0.5/0.8, 0.8/1.1, 1/1.3, 1.3/1.6, 1.6/1.8, 1.9/2.1, 2.1/2.3, 2.4/2.5, 2.7/2.7, 3/3, 3.2/3.2, 3.5/3.4, 
                   3.8/3.6, 4/3.8, 4.3/4, 4.6/4.2, 4.9/4.5} {
    \fill[orange] (\x,\y) circle (3pt);
}

\node[anchor=west] at (4.75,5.5) {Sharp bounding function};
\node[anchor=west] at (4.5,5) {$y = mP_j + b$};

\node[anchor=west] at (5.25,3.0) {\textbf{CONJECTURE:}};
\node[align=left] at (7.5,2) {
    If $\mathcal{O}$ is an object and $\mathcal{O} \in H_1$, then \\ $P_i(\mathcal{O}) \leq mP_j(\mathcal{O}) + b$,\\
    and this bound is sharp.
};

\end{tikzpicture}
\caption{Finding a possible (linear) upper bound on the target property $P_i$ in terms of property $P_j$ for objects in $H_1$. Data points associated with containment in $H_1$ shown by orange dots, whereas instances no in $H_1$ shown by green dots. }
\label{fig:main2}
\end{figure}
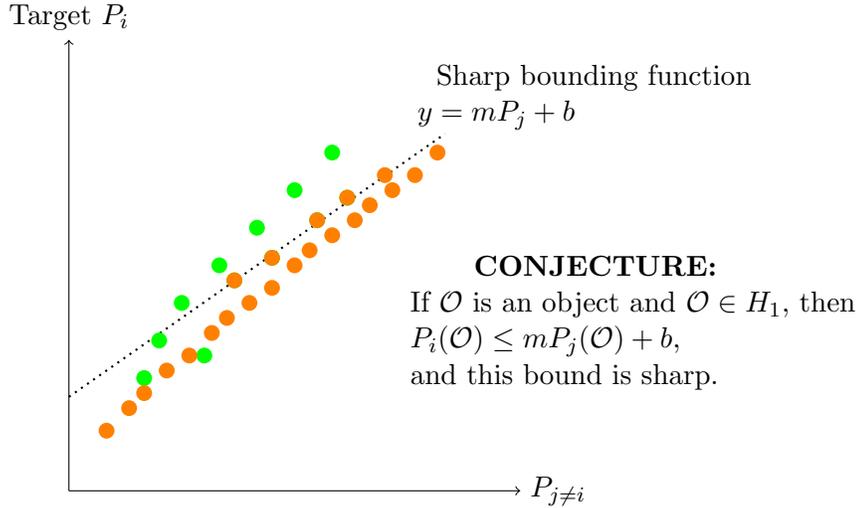
The above process is done on each numerical column of the feature data; that is, each pair of numerical functions is compared against each other and given a proposed inequality conjectured between them. Each conjecture generated by the above steps applies to all types of objects in the database. However, we can generate even more conjectures. By applying the same steps to a subset of objects in the database that satisfy a particular Boolean property (or combination of Boolean properties), we may obtain conjectures that are less general but potentially stronger; see Figure~\ref{fig:main2} for an illustration of this process.

\subsection{Sorting and Filtering}\label{subsec:filter}

Once the optimal bounding functions for a given target invariant are found, we are left with a list of possible conjectures, along with detailed data for each conjecture. This data includes the set of objects that satisfy the conjecture's hypothesis, the graphs that attain equality, and the count of these graphs. At this point, our program implements a sorting procedure. That is, the list of conjectures is then sorted in nonincreasing order with respect to the touch number of the conjectures. Thus, the conjectures at the top of the list hold with equality more than the conjectures towards the bottom of the list. Thus, this sorting aspect implemented by our program insures the conjectures at the top of the list are \emph{``stronger''} than those at the bottom of the list. 

After the conjectures have been sorted according to their respective touch numbers, our program then implements the first of two filtering heuristics, called \emph{Theo}. This heuristic checks if any proposed inequality relation appears more than once in the list of conjectures, and then only selects the proposed conjectures with this inequality that have the most general hypothesis statement. For example, consider the following two conjectures. 
\begin{conj}\label{conj:alpha-mu-cubic1}
If $G$ is a connected and cubic graph, then
\[
\alpha(G) \leq \mu(G),
\]
and this bound is sharp. 
\end{conj}

\begin{conj}\label{conj:alpha-mu-reg1}
If $G$ is a connected and $r$-regular graph with $r > 0$, then
\[
\alpha(G) \leq \mu(G),
\]
and this bound is sharp. 
\end{conj}
The Theo heuristic would automatically detect that the inequality $\alpha(G) \leq \mu(G)$ appears in both Conjecture~\ref{conj:alpha-mu-cubic1} and in Conjecture~\ref{conj:alpha-mu-reg1}, and thereafter, check if the set of graphs in the database that satisfy the hypothesis of Conjecture~\ref{conj:alpha-mu-cubic1} also satisfy the hypothesis of Conjecture~\ref{conj:alpha-mu-reg1}. Since every cubic graph is also a regular graph, but not vice versa, Theo would remove Conjecture~\ref{conj:alpha-mu-cubic1} from the possible conjectures to present to the user. The reasoning for this heuristic is to put more emphasize on the more general conjecture, and thus, remove any conjecture that may follow from a more general statement.

The list of conjectures is (optionally) further filtered by a variation of the Dalmation heuristic which we call \emph{Dalmation-static}. Unlike the original Dalmation heuristic used by GRAFFITI~\cite{Larson}, our version of Dalmation takes as input a \emph{static list} of conjectures and works as following:
\\

\noindent\textbf{Dalmation-Static Heuristic:}
\begin{enumerate}
    \item[1.] Let $\mathcal{G}$ be the set of graphs attaining equality in Conjecture 1 in the current list of conjectures, recalling that the conjectures are sorted according to their respective touch number. 
    \item[2.] For $i = 2, \dots, N$, if the set of graphs attaining equality in Conjecture $i$ does not contain a which is also graph not contained in $\mathcal{G}$, then remove Conjecture $i$ from the list of conjectures, otherwise, let $\mathcal{G} = \mathcal{G} \cup \mathcal{G}_i$, where $\mathcal{G}_i$ is the set of graphs attaining equality in Conjecture $i$. 
\end{enumerate}

Finally, the conjectures are further filtered by removing any known conjectures. This aspect of TxGraffiti requires maintenance and is one area where the program would benefit from many mathematicians contributing to. Moreover, since conjectures are computed at each prompt to the program, anytime a new counter-example is added to the database, new conjectures appear. This aspect motivated further development of the program, where users may enter in counter-examples to better the conjectures of the program. This functionality is available by emailing the author via the online website, but will in the future allow for a more straight-forward approach. 

The resulting set of conjectures is a set of conjectures that can be viewed as \emph{``mathematical strong''}. That is, the linear optimization methods first find proposed inequalities that are guaranteed to be sharp on a maximum number of graph instances, thereafter, the Theo heuristic insures generality of the hypothesis for a conjectures inequality, then the Dalmation-static heuristic insures  presented conjectures only consist of inequalities providing \emph{``new''} information, and finally the touch number sorting insures conjectures at the beginning of a list are sharper than ones that follow. In the following section we demonstrate how these processes show promise for new mathematical insight in the realm of graph theory. 

\subsection{Code and Reproducibility}\label{subsec:code}
For readily available examples of this process, see the GitHub repository associated with the interactive website~\cite{TxGraffiti}.

\section{Results}\label{sec:results}
In this section we present some results stimulated by conjectures of TxGraffiti. More specifically, we highlight the following results inspired by conjectures of TxGraffiti and listed in Table~\ref{tab:graph-theory-conjectures}. Of the results listed in Table~\ref{tab:graph-theory-conjectures}, we now focus on the result pertaining to the \emph{independence number} and \emph{matching number} of regular graphs. The originial conjecture that stimulated this result states that for any 3-regular and connected graph $G$, the independence number $\alpha(G)$ is at most the matching number $\mu(G)$.  
\begin{conj}\label{conj:alpha-mu}
If $G$ is a connected and cubic (3-regular) graph, then 
\[
\alpha(G) \leq \mu(G),
\]
where $\alpha(G)$ is the independence number and $\mu(G)$ is the matching number.
\end{conj}

\begin{table}[h]
\centering
\begin{tabular}{|c|c|c|}
\hline
\textbf{Conjecture} & \textbf{Graph Family} & \textbf{Authors and Publication} \\
\hline
$\alpha(G) \leq \mu(G)$ & cubic graphs & Caro et al.~\cite{CaDaPe2020}\\
\hline
$Z(G) \leq \beta(G)$ & claw-free graphs & Brimkov et al.~\cite{TxGraffiti-2023} \\
\hline
$\alpha(G) \leq \frac{3}{2}\gamma_t(G)$ & cubic graphs & Caro et al.~\cite{CaDaHePe2022} \\
\hline
$\alpha(G) \leq \gamma_2(G)$ & claw-free graphs & Caro et al.~\cite{CaDaHePe2022} \\
\hline
$\gamma_e(G) \geq \frac{3}{5}\mu(G)$ & cubic graphs & Caro et al.~\cite{CaDaHePe2022} \\
\hline
$Z(G) \leq 2\gamma(G)$ & cubic graphs & Davila and Henning~\cite{DaHe21a}\\
\hline
$Z_t(G) \leq \frac{3}{2}\gamma_t(G)$ & cubic graphs & Davila and Henning~\cite{DaHe19b}  \\
\hline
$Z(G) \leq \gamma(G) + 2$ & cubic claw-free graphs & Davila~\cite{Davilazdom}  \\
\hline
\end{tabular}
\caption{Notable conjectures in graph theory generated by TxGraffiti and their corresponding publications.}
\label{tab:graph-theory-conjectures}
\end{table}

Notably, Conjecture~\ref{conj:alpha-mu} relates three of the oldest studied properties in graph theory; namely, independent sets, matching sets, and regular graphs. For this reason, the author did not share Conjecture~\ref{conj:alpha-mu} for many months, believing it to be trivial known. However, once shared with collaborators, it became apparent that not only was this conjecture not known in the literature, but was also true. Indeed, this conjecture was then generalized and proven, resulting in the following theorem; the proof of which appears in~\cite{CaDaPe2020}, but is also given below to demonstrate this simple and meaningful result. 

\begin{thm}[Caro et al.~\cite{CaDaPe2020}]\label{thm:alpha-mu}
If $G$ is an $r$-regular graph with $r > 0$, then 
\[
\alpha(G) \leq \mu(G),
\]
and this bound is sharp.
\end{thm}
\proof Let $G$ be an $r$-regular graph with $r > 0$. Let $X \subseteq V(G)$ be a maximum independent set, and let $Y = V(G) \setminus X$. By removing edges from $G$ that have both endpoints in $Y$, we form a bipartite graph $H$ with partite sets $X$ and $Y$.

Since the removed edges were only those with both endpoints in $Y$, any vertex in $X$ will have the same open neighborhood in $H$ as it does in $G$. Given that $G$ is $r$-regular and $X$ is an independent set, each vertex in $X$ will have exactly $r$ neighbors in $Y$.

Let $S \subseteq X$ be chosen arbitrarily, and let $e(S, N_H(S))$ denote the number of edges from $S$ to $N_H(S)$. Since each vertex in $S$ has exactly $r$ neighbors in $Y$, it follows that $e(S, N_H(S)) = r|S|$. Additionally, since each vertex in $N_H(S)$ has at most $r$ neighbors in $X$, we also have $e(S, N_H(S)) \leq r|N_H(S)|$. Thus, $r|S| \leq r|N_H(S)|$, implying that $|S| \leq |N_H(S)|$. By Hall’s Theorem; see West~\cite{West}, there exists a matching $M$ that can match $X$ to a subset of $Y$. Since $X$ is a maximum independent set and $M$ is also a matching in $G$, we conclude that $\alpha(G) = |M| \leq \mu(G)$, proving the theorem.
\qed 

Notably, by confirming Conjecture~\ref{conj:alpha-mu-cubic1} with Theorem~\ref{thm:alpha-mu}, we were inspired to include the more general hypothesis of regular graphs in conjectures presented by TxGraffiti, and this resulted in the more general statement of Theorem~\ref{thm:alpha-mu} being presented as a conjecture by the program. From an application point of view, the resulting theorem due to Conjecture~\ref{conj:alpha-mu-cubic1} is further interesting since the computation of the matching number $\mu(G)$ is computable in polynomial time, whereas the computation of the independent number $\alpha(G)$ is NP-hard. Thus, the resulting theory gathered from investigating Conjecture~\ref{conj:alpha-mu-cubic1} has practical applications in graph theory and the sciences.

\section{Conclusion}\label{sec:conclusion}

In this paper, we have described the artificial intelligence program TxGraffiti and also provided evidence for its usefulness in mathematical research as its conjectures span various areas of graph theory; many leading to significant publications. Moreover, we provide a new web-based interaction for TxGraffiti which may lead to even further mathematical insight. We anticipate that further development and application of the ideas underpinning TxGraffiti will lead to further insights into computer assisted mathematics.

\end{document}